\RequirePackage{fix-cm}
\documentclass[twocolumn]{svjour3-sa}          % twocolumn
\smartqed  % flush right qed marks, e.g. at end of proof
\usepackage{graphicx}
\RequirePackage{mmap}
\usepackage{amsmath}
\usepackage{amsfonts}
\usepackage{amssymb}
\usepackage{graphicx}
\usepackage{subfigure}
\usepackage{url}
\usepackage{mathtools}
\usepackage{floatrow}

\begin{document}

\title{Multi-Scale Anisotropic Fourth-Order Diffusion Improves Ridge
  and Valley Localization%\thanks{Grants or other notes
}


\author{Shekoufeh Gorgi Zadeh    \and
		Stephan Didas \and
        Maximilian W. M. Wintergerst \and %etc.
        Thomas Schultz
}


\institute{
			Shekoufeh Gorgi Zadeh \and Thomas Schultz
			\at
            Institute of Computer Science II, University of Bonn, Germany \\
            \email{gorgi@cs.uni-bonn.de}           %  \\
			\and 
			Stephan Didas
			\at Umwelt-Campus Birkenfeld, Hochschule Trier, Germany
			\and
			Maximilian W. M. Wintergerst
			\at Department of Ophthalmology, University of Bonn, Germany
}

\date{Received: 25 November 2016 / Accepted: 30 March 2017}

\maketitle

\begin{abstract}
Ridge and valley enhancing filters are widely used in applications
  such as vessel detection in medical image computing. When images are
  degraded by noise or include vessels at different scales, such
  filters are an essential step for meaningful and stable vessel
  localization. In this work, we propose a novel multi-scale
  anisotropic fourth-order diffusion equation that allows us to smooth
  along vessels, while sharpening them in the orthogonal direction.
  The proposed filter uses a fourth order diffusion tensor whose
  eigentensors and eigenvalues are determined from the local Hessian
  matrix, at a scale that is automatically selected for each pixel. We
  discuss efficient implementation using a Fast Explicit Diffusion
  scheme and demonstrate results on synthetic images and vessels in
  fundus images. Compared to previous isotropic and anisotropic
  fourth-order filters, as well as established second-order vessel
  enhancing filters, our newly proposed one better restores the
  centerlines in all cases.
\end{abstract}

\section{Introduction}
\label{intro}

In image analysis, ridges and valleys are curves along which the image
is brighter or darker, respectively, than the local background
\cite{eberly1994ridge}. Collectively, ridges and valleys are referred
to as creases. Reliable detection and localization of creases in noisy
images is an important and well-studied problem in medical image
analysis, one very common application being the detection of blood
vessels \cite{frangi1998multiscale}.

Often, ridges and valleys occur at multiple scales, i.e., their
cross-sectional radius varies throughout the image. For example, the
stem of a vessel tree is thicker than its branches. Gaussian scale
spaces are a classic strategy for extracting creases at different
scales \cite{lindeberg1998edge}. However, the fact that Gaussian
filters do not offer any specific mechanisms for preserving creases
gave rise to image filters such as coherence enhancing diffusion
\cite{weickert1998anisotropic}, crease enhancement diffusion (CED)
\cite{sole2001crease}, and vesselness enhancement diffusion (VED)
\cite{canero2003vesselness}. They are based on second order
anisotropic diffusion equations with a diffusion tensor that, in the
presence of crease lines, smoothes only along, but not across them. In
addition, the VED filter includes a multi-scale analysis that
automatically adapts it to the local scale of creases.

In this work, we argue that using fourth-order instead of second-order
diffusion to enhance creases allows for a more accurate localization
of their centerlines. We propose a novel fourth-order filter that
introduces a fourth-order diffusion tensor to specifically enhance
ridges, valleys, or both, in a scale-adaptive manner. Increased
accuracy of the final segmentation is demonstrated on simulated and
real-world medical images.

\section{Related Work}
\label{sec:related-work}

Diffusion-based image filters treat image intensities as an initial
heat distribution $u_{t=0}$, and solve the heat equation
$\partial_t u=\mathrm{div} (g\,\nabla_{\mathbf{x}} u)$ for larger
values of an artificial time parameter $t$, corresponding to
increasingly smoothed versions of the image. If the diffusivity
function $g$ is constant, the diffusion is linear and uniformly
smoothes image $u$.  If $g=1$, the solution at time
$t$ can be obtained as the convolution
$u*G_{\sigma}$ with a Gaussian kernel
$G_{\sigma}$ with standard deviation $\sigma=\sqrt{2t}$
\cite{weickert1998anisotropic}.
 
Since linear diffusion fails to preserve important image structures,
Perona and Malik \cite{perona1990scale} introduced the idea of using
\emph{nonlinear} diffusion equations. By making the scalar diffusivity
$g$ a function of the spatial gradient magnitude
$\|\nabla_{\mathbf{x}} u\|$, they reduce the amount of smoothing near
image edges, and thus preserve edges. One such diffusivity function is
\begin{equation}
\label{eq:perona-malik}
g\left(\left\|\nabla_{\mathbf{x}} u\right\|^{2}\right)=\frac{1}{1+\frac{\left\|\nabla_{\mathbf{x}} u\right\|^{2}}{\lambda^{2}}} \text{ ,}
\end{equation}
where $\lambda$ is called the contrast parameter, and determines the
minimum strength of edges that should be preserved
\cite{perona1990scale}.

\begin{figure}
\centering
\begin{tabular}{@{}c@{ }c@{ }c@{}}

% &  & \tabularnewline
\includegraphics[scale=0.081]{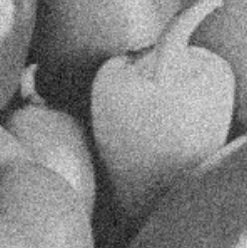} &\includegraphics[scale=0.081]{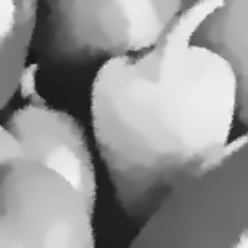}&\includegraphics[scale=0.081]{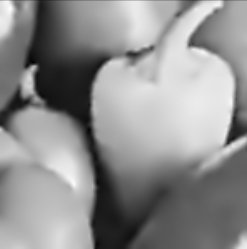}\tabularnewline
\shortstack{Input \\ \quad \\ \quad} & \shortstack{Second-order \\ diffusion} & \shortstack{Fourth-order \\ diffusion}\tabularnewline
\end{tabular}\caption{On smoothly shaded surfaces, second-order Perona-Malik diffusion creates a staircasing artifact that is avoided by fourth-order diffusion. \label{fig:staircase-artifact}}
\end{figure}

Perona-Malik diffusion turns smoothly shaded surfaces into piecewise
constant profiles, an effect that is often referred to as a
staircasing artifact, and that can be seen in the central ``Pepper''
image in Figure~\ref{fig:staircase-artifact}. To avoid this effect,
higher-order diffusion replaces the two first-order spatial
derivatives in the heat equation with second-order derivatives. More
recently, higher-order PDEs were also generalized to implicit surfaces
\cite{greer2006fourth} and image colorization \cite{peter2016turning}.

In a one-dimensional setting, discrete variants of higher order data
regularization can be traced back to a 1922 article by Whittaker
\cite{Whi22}.  A first approach for higher order regularization in
image processing involving the absolute value of all second order
partial derivatives has been proposed by Scherzer \cite{Sch98}. The
resulting method has the drawback of not being rotationally invariant.
An extension of classical regularization by choosing $(\Delta u)^2$ as
argument of the penalizer in the smoothness term has been proposed by
You and Kaveh \cite{YK00}. Their method introduces speckle artifacts
around edges that require some post-processing.  Both problems can be
solved by using the squared Frobenius norm of the Hessian matrix
$\| H(u) \|^2_F$ as argument of the penalizer. This has been proposed
by Lysaker et al.\ \cite{lysaker2003noise}.

Two very similar higher order methods based on evolution equations without underlying variational formulation have been introduced by Tumblin and Turk \cite{TT99} and Wei \cite{Wei99}. They use fourth-order evolution equations of the form 
\begin{equation}
\partial_t u \; = \; - \mbox{div} \Big( g(m) \nabla \Delta u \Big) \text{,}
\end{equation}
where $m$ is the gradient norm \cite{TT99} or the Frobenius norm of the Hessian \cite{Wei99}. 

Didas et al.\ \cite{didas2009properties} have generalized higher order
regularization methods and the corresponding partial differential
equations to arbitrary derivative orders. They have
shown that, when combined with specific diffusivity functions,
\emph{fourth-order} equations can enhance image curvature analogous to
how, by careful use of forward and backward diffusion, second-order
equations can enhance, rather than just preserve, image edges
\cite{weickert1998anisotropic}.

\begin{figure}
  \centering\includegraphics[scale=0.26]{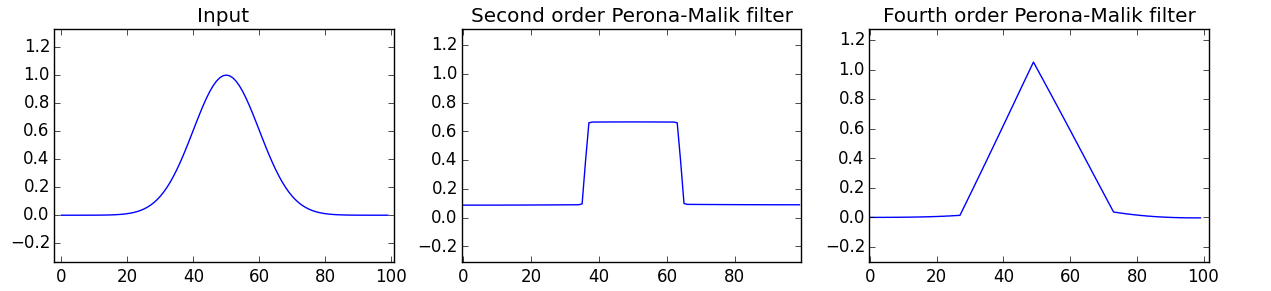}\protect\caption{
    Second-order diffusion filter and fourth-order diffusion filter
    applied on a 1-D input signal.  Both filters use the Perona-Malik
    diffusivity function.\label{fig:second-vs-fourth-order-diffusion}}
\end{figure}

Figure~\ref{fig:second-vs-fourth-order-diffusion} shows a simple
one-dimensional example with the Perona-Malik diffusivity function
(\ref{eq:perona-malik}). The edge enhancement of nonlinear
second-order diffusion, which leads to a piecewise constant result,
and the curvature enhancement of nonlinear fourth-order diffusion,
which leads to a piecewise linear result, are clearly
visible. Obviously, localizing the maximum will be much easier and
more reliable in case of the sharp peak created by fourth-order
diffusion than in the extended plateau that results from second-order
diffusion.

It is this curvature-enhancing property of fourth-order diffusion that
we exploit in our novel filter. We combine it with the idea of
\emph{anisotropic} diffusion, which was introduced to image processing
by Weickert \cite{weickert1998anisotropic} to address another
limitation of the Perona-Malik model. Namely, a consequence of
preserving edges by locally reducing the amount of smoothing is that
the neighborhoods of edges remain noisy. Anisotropic diffusion
$\partial_t u=\mathrm{div} (D\,\nabla_{\mathbf{x}} u)$ replaces the
scalar diffusivity $g$ by a second-order diffusion tensor $D$, which
makes it possible to reduce smoothing orthogonal to, but not along
image features, and therefore to denoise edges more effectively than
isotropic nonlinear diffusion, while still avoiding to destroy them.

We generalize this approach to a novel anisotropic fourth-order
diffusion equation that smoothes along the crease, while creating a
sharp peak in the orthogonal direction, to clearly indicate its
center. We are aware of only one previous formulation of anisotropic
fourth order diffusion, proposed by Hajiaboli
\cite{hajiaboli2011anisotropic}. However, it has been designed to
preserve edges, rather than enhance creases. Consequently, it is not
well-suited for our purposes, as we will demonstrate in the
results. Moreover, it differs from our approach in that it does not
make use of a fourth-order diffusion tensor, and includes no mechanism
for scale selection.

Despite the long history of research in this area, improved filtering
and detection of ridges continues to be an active topic in medical
image analysis. Our work on improving localization through
fourth-order diffusion complements recent advances. For example, the
SCIRD ridge detector by Annunziata et al.\ \cite{annunziata2015scale},
or the vesselness measure by Jerman et al.\
\cite{jerman2016enhancement} that gives better responses for vessels
of varying contrasts, could replace the vessel segmentation by Frangi
et al.\ \cite{frangi1998multiscale} that we use as a prefiltering
step. Several recent works
\cite{franken2009crossing,hannink2014crossing,scharr2012short,stuke2004analysing}
have addressed diffusion in crossings and bifurcations, and could be
combined with our work to improve the performance of our filter in
such cases.

%Fourth-order diffusion generalizes the heat equation by replacing all
%first-order spatial derivatives with second-order derivatives. 
%Our new filter is based on a nonlinear fourth-order diffusion equation that
%was introduced by Lysaker et al. \cite{lysaker2003noise} with the
%primary goal of avoiding the staircasing artifacts that occur in
%second-order Perona-Malik diffusion \cite{perona1990scale}, as shown on
%the ``Pepper'' image in Figure~\ref{fig:staircase-artifact}.  Subsequently, Didas et
%al. \cite{didas2009properties} have shown that, when combined with
%specific diffusivity functions, such fourth-order equations can
%enhance image curvature analogous to how, by careful use of forward
%and backward diffusion, second-order equations can enhance, rather
%than just preserve, image edges \cite{weickert1998anisotropic}.

\section{Method}

\subsection{Anisotropic Fourth-order Diffusion}

%Diffusion-based image filters treat image intensities as an initial
%heat distribution $u_{t=0}$, and solve the heat equation $\partial_t
%u=\mathrm{div} (g\,\nabla_{\mathbf{x}} u)$ for larger values of an
%artificial time parameter $t$, corresponding to increasingly smoothed
%versions of the image. When the scalar diffusivity $g$ is made a
%function of the spatial gradient magnitude $\|\nabla_{\mathbf{x}}
%u\|$, diffusion becomes nonlinear, and image edges can be preserved by
%reducing the amount of smoothing near them \cite{perona1990scale}.

%\emph{Anisotropic} diffusion
%$\partial_t u=\mathrm{div} (D\,\nabla_{\mathbf{x}} u)$ replaces the
%scalar diffusivity $g$ by a second-order diffusion tensor $D$, which
%makes it possible to reduce smoothing orthogonal to, but not along
%image features, and therefore to denoise edges more effectively than
%isotropic nonlinear diffusion, while still avoiding to destroy them
%\cite{weickert1998anisotropic}. 

% As already mentioned in Section~\ref{sec:related-work},
% fourth-order filters generalize
% the diffusion equation in a different way, by replacing the two
% first-order spatial derivatives in the heat equation with second-order
% derivatives, leading to an overall order of four. In particular,

Building on work of Lysaker et al.\ \cite{lysaker2003noise}, Didas et
al.\ \cite{didas2009properties} formulate nonlinear fourth-order
diffusion as
\begin{equation}
\label{eq:nonlinear-fourth-order}
\begin{aligned}
\partial_t u = &-\partial_{xx}(g(\|H(u)\|_F^2)u_{xx})
-\partial_{yx}(g(\|H(u)\|_F^2)u_{xy})\\
&-\partial_{xy}(g(\|H(u)\|_F^2)u_{yx})
-\partial_{yy}(g(\|H(u)\|_F^2)u_{yy})\text{,}
\end{aligned}
\end{equation}
where $\|H(u)\|_F^2$ is the Frobenius norm of the Hessian matrix of image $u$, and 
$u_{xy}=\partial_{xy} u$.
We propose the following novel \emph{anisotropic fourth-order} diffusion
model, which combines the ideas of higher-order diffusion with that of
making diffusivity a function of both spatial location and direction:
\begin{equation}
\label{eq:ts-fourth-order-diffusion}
\begin{aligned}
\partial_{t}u = &-\partial_{xx}\left[\mathcal{D}(H_{\rho}(u_{\sigma})):H(u)\right]_{xx}\\
                &-\partial_{yx}\left[\mathcal{D}(H_{\rho}(u_{\sigma})):H(u)\right]_{xy}\\
                &-\partial_{xy}\left[\mathcal{D}(H_{\rho}(u_{\sigma})):H(u)\right]_{yx}\\
                &-\partial_{yy}\left[\mathcal{D}(H_{\rho}(u_{\sigma})):H(u)\right]_{yy}\text{.}
\end{aligned}
\end{equation}

Equation~(\ref{eq:ts-fourth-order-diffusion}) introduces a general
linear map $\mathcal{D}$ from the Hessian matrix $H$ to a transformed
matrix. Linear maps from matrices to matrices are naturally written as
fourth-order tensors, and we use the ``double dot product''
$\mathcal{D}:H$ as a shorthand for applying the map $\mathcal{D}$ to
the Hessian matrix $H$. This results in a transformed matrix $T$, and
we use square brackets $[T]_{ij}$ to denote its $(i,j)$th
component. Formally,
\begin{equation}
  \label{eq:double-contraction}
  \begin{aligned}
  \left[T\right]_{ij} & =  \left[
  \mathcal{D}(H_{\rho}(u_{\sigma})):H(u)\right]_{ij}\\   & =  \sum_{k=1}^2\sum_{l=1}^2\left[\mathcal{D}(H_{\rho}(u_{\sigma}))\right]_{ijkl}\left[H(u)\right]_{kl}.
  \end{aligned}
\end{equation}

In this notation, we can define second-order eigentensors $E$ of
$\mathcal{D}$ corresponding to eigenvalue $\mu$ by the equation
$\mathcal{D}:E=\mu E$. An alternative notation, which will be used for
the numerical implementation in Section~\ref{sec:l2-stability}, writes
the Hessian and transformed matrices as vectors. This turns
$\mathcal{D}$ into a matrix whose eigenvectors are nothing but the
vectorized eigentensors as defined above. Similar to others
\cite{Basser:2007,Kindlmann:2007}, we find the fourth-order tensor and
``double dot'' notation more appealing for reasoning at a higher
level, because it allows us to preserve the natural structure of the
involved matrices.

Using our square bracket notation, an equivalent way of writing one of
the terms from Equation~(\ref{eq:nonlinear-fourth-order}),
$\partial_{ji}(g(\|H(u)\|_F^2)u_{ij})$, is
$\partial_{ji}\left[g(\|H(u)\|_F^2)H(u)\right]_{ij}$. Thus, the
difference between the model from
Equation~(\ref{eq:nonlinear-fourth-order}) and our new one in
Equation~(\ref{eq:ts-fourth-order-diffusion}) is to replace the
isotropic scaling of Hessian matrices using a scalar diffusivity
$g$, with a general linear transformation $\mathcal{D}$, which acts on
the second-order Hessian in analogy to how the established
second-order diffusion tensor acts on gradients in second-order
anisotropic diffusion. Due to this analogy, we call $\mathcal{D}$ a
fourth-order diffusion tensor.

In our filter, $\mathcal{D}$ is a function of the local normalized
Hessians, which are defined as
\begin{equation}
  \label{eq:local_normalized_hessian}
  H_{\rho}(u_{\sigma})=G_{\rho}\ast\left(\frac{1}{\sqrt{1+\left\Vert \nabla u_{\sigma}\right\Vert }}H\left(u_{\sigma}\right)\right)\text{, }
\end{equation}
where regularized derivatives are obtained by convolution with a
Gaussian kernel, $u_{\sigma}:=u\ast G_{\sigma}$. Its width $\sigma$
should reflect the scale of the crease, as will be discussed in
Section~\ref{sec:scale-selection}. Since scale selection might
introduce spatial discontinuities in the chosen $\sigma$, the
normalized Hessians are made differentiable by integrating them over a
neighborhood, for which we use a Gaussian width $\rho=0.5$ in our
experiments.  As shown in \cite{haralick:1983}, and used for
vesselness enhancement diffusion in \cite{canero2003vesselness}, the
inverse gradient magnitude factor is used to make the eigenvalues of
$H_{\rho}(u_{\sigma})$ match the surface curvature values.

%Parameter $\rho$ is to make the
%resulting scale image differentiable, we perform Gaussian smoothing
%with $\rho=0.5$.

We emphasize that, unlike in a previous generalization of structure
tensors to higher order \cite{schultz2009higher}, the reason for going
to higher tensor order in
Equation~(\ref{eq:ts-fourth-order-diffusion}) is not to preserve
information at crossings; this is a separate issue that was recently
addressed by others \cite{hannink2014crossing}, and that we plan to
tackle in our own future work. In our present work, our goal is to
smooth along ridges and valleys, while sharpening them in the
orthogonal direction. This sharpening requires the curvature-enhancing
properties of fourth-order diffusion, and a fourth-order diffusion
tensor is a natural consequence of making fourth-order diffusion
anisotropic.

\subsection{Fourth-order Diffusion Tensor $\mathcal{D}$}
\label{sec:fourth-order-diff-tensor-D}

We now need to construct our fourth-order diffusion tensor
$\mathcal{D}$ so that it will smooth along creases, while enhancing
them in the perpendicular direction. Similar to Weickert's diffusion
tensors \cite{weickert1998anisotropic}, we will construct
$\mathcal{D}$ in terms of its eigentensors $E_i$ and corresponding
eigenvalues $\mu_i$, as defined above.

Didas et al.\ \cite{didas2009properties} have shown that fourth-order
diffusion with the Perona-Malik diffusivity \cite{perona1990scale}
allows for adaptive smoothing or sharpening of image curvature,
depending on a contrast parameter $\lambda$. In particular, in the 1-D
case, only forward diffusion (i.e., smoothing) happens in regions with
$|\partial_{xx}u|<\lambda$, while only backward diffusion (i.e.,
curvature enhancement) occurs where
$|\partial_{xx}u|>\sqrt{3}\lambda$. We wish to exploit this to enhance
creases whose curvature is strong enough to begin with, while
smoothing out less significant image features.

This is achieved by deriving the eigenvalues $\mu_{i}$ of
$\mathcal{D}$ from the eigenvalues $\nu_{1},\nu_{2}$ of the normalized
Hessian $H_{\rho}(u_{\sigma})$ using the Perona-Malik diffusivity
\cite{perona1990scale}, i.e.,
\begin{equation}
\mu_{i}=\frac{1}{1+\nu_{i}^{2}/\lambda^{2}} \text{, for }i\in\{1,2\}\label{eq:eigenvalues-in-2d}\text{.}
\end{equation}

If the user wishes to specifically enhance either ridges or valleys,
the sign of $\nu_{i}$ could be taken into account. For instance, a
ridge-like behaviour in the $i$th direction is characterized by
$\nu_{i}<0$. Therefore, we can decide to smooth out valleys by setting
$\mu_{i}=1$ wherever $\nu_{i}\geq 0$, and enhance ridges wherever
$\nu_{i}<0$ by defining $\mu_{i}$ as before. Enhancing only valleys
can be done in full analogy. In our experiments on synthetic data, we
found that, in terms of the $\ell_{2}$ difference between the ground
truth and the filtered image, better results were obtained when
enhancing both ridges and valleys. This is the setting used in all our
experiments.

The ridge and valley directions can be found from the eigenvectors
$e_{1},e_{2}$ of the normalized Hessian matrix $H_{\rho}(u_{\sigma})$, and are reflected
in the eigentensors $E_{i}$ of $\mathcal{D}$ by setting
\begin{equation}
\begin{array}{cccc}
E_{1}= & e_{1}\otimes e_{1}\qquad\quad & E_{3}= & \frac{1}{\sqrt{2}}(e_{1}\otimes e_{2}+e_{2}\otimes e_{1})\\
E_{2}= & e_{2}\otimes e_{2}\qquad\quad & E_{4}= & \frac{1}{\sqrt{2}}(e_{1}\otimes e_{2}-e_{2}\otimes e_{1})
\end{array}\text{ .}\label{eq:eigentensors-in-2d}
\end{equation}

The $E_i$ are orthonormal with respect to the tensor dot product
$A:B:=\mathrm{tr}(B^\mathrm{T}A)$. By definition, $E_4$ is
antisymmetric. Since Hessians of smooth functions are symmetric, the
value of $\mu_{4}$ does not play a role, and is simply set to zero. We
define $\mu_{3}$ as the average of $\mu_{1}$ and $\mu_{2}$.

%$E_{3}$ is a symmetric tensor whereas $E_{4}$ is an antisymmetric
%one.
% The outer product of the Hessian matrix with itself could also be used
% to define the fourth order diffusion tensor $\mathcal{D}$ but with this
% method the sign of individual eigenvalues cannot be correctly reconstructed
% and therefore the option of enhancing only either ridges or valleys is no longer there.
 
\subsection{Scale Selection}
\label{sec:scale-selection}

In the previous sections, crease orientation was estimated using the
eigenvectors of the regularized and normalized Hessian in
Equation~(\ref{eq:local_normalized_hessian}). As in previous
approaches such as vesselness enhancement diffusion (VED)
\cite{canero2003vesselness}, this involves a regularization parameter
$\sigma$ that should be adapted to the local radius of the
crease. Setting this parameter is referred to as scale selection.

The vesselness measure $\mathcal{V}_{\sigma}$ by Frangi et
al. \cite{frangi1998multiscale} is maximal at the scale $\sigma$ that
matches the corresponding vessel size, and has been widely used for
detecting the local radius of vessel like
structures. $\mathcal{V}_{\sigma}$ is obtained from sorted and
scale-normalized eigenvalues $|\tilde\nu_{1}|\leq|\tilde\nu_{2}|$,
computed as $\tilde\nu_i:=\sigma^2 \nu_i$ from eigenvalues $\nu_i$ of
the Hessian $H(u_{\sigma})$ at a given scale $\sigma$. The factor
$\sigma^2$ compensates for the loss of contrast at larger scales
\cite{lindeberg1998edge}.

A vesselness measure $\mathcal{V}_{\sigma}$ should be low in
background regions where overall curvature and thus
$\mathcal{S}=\sqrt{\tilde\nu_{1}^{2}+\tilde\nu_{2}^{2}}$ are low
overall. Moreover, it should detect tubular structures, where
$|\tilde\nu_{1}|\ll|\tilde\nu_{2}|$, as opposed to blobs, in which
$\mathcal{R_{B}=}\frac{\tilde\nu_{1}}{\tilde\nu_{2}}$ would be
large. For ridges ($\tilde\nu_2<0$), Frangi et al.\ achieve this by
combining $\mathcal{S}$ and $\mathcal{R_{B}}$ according to
\begin{equation}
\mathcal{V}_{\sigma}u=\begin{cases}
0 & \text{\quad if }\quad\tilde\nu_{2}>0\\
\left(\text{e}^{-\frac{\mathcal{R_{B}^{\mathrm{2}}}}{2\beta^{2}}}\right)\left(1-\text{e}^{-\frac{\mathcal{S}^{2}}{2c^{2}}}\right) & \text{\quad otherwise}
\end{cases}\text{ ,}\label{eq:vesselness-measure}
\end{equation}
where the $\beta$ and $c$ parameters tune $\mathcal{V}_{\sigma}u$ to
be more specific with respect to suppression of blob shapes or
background structures, respectively. We use $\beta=0.5$ and
$c=\frac{1}{2}(\text{max}(\mathcal{S}))$, as recommended in \cite{frangi1998multiscale}.

The scale for each pixel is selected as the $\sigma$ for which the
maximum
$\mathcal{V}u=\max_{\sigma=\sigma_{\text{min}},...,\sigma_{\text{max}}}\mbox{\ensuremath{\mathcal{V}_{\sigma}}\ensuremath{u}}$
is attained, where $\{\sigma_{\text{min}},...,\sigma_{\text{max}}\}$
are the range of expected scales in the image.  For pixels that are
part of the background, $\mathcal{V}u$ is low, and it can be
thresholded by parameter $\theta\in\left[0,1\right]$ for vessel
segmentation. This segmentation indicates the extent of vessels, and
is used for our scale-image postprocessing, as described below.

\begin{figure}
\centering%
\begin{tabular}{c@{ }c@{ }c@{}c}
Input & Scale Image & \shortstack{Post Processed} & \tabularnewline
\includegraphics[scale=0.084]{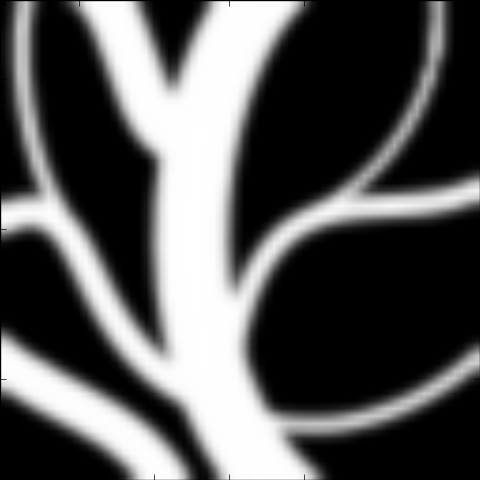} & \includegraphics[scale=0.084]{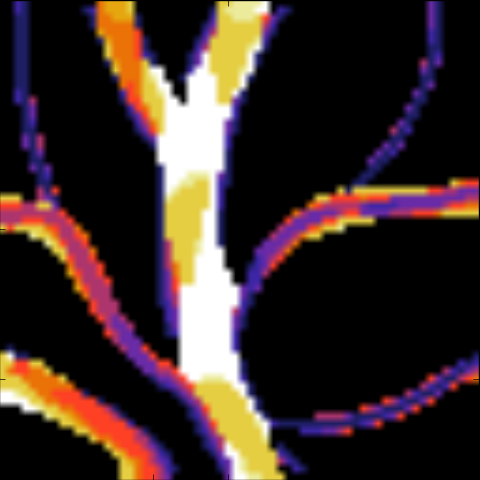} & \includegraphics[scale=0.084]{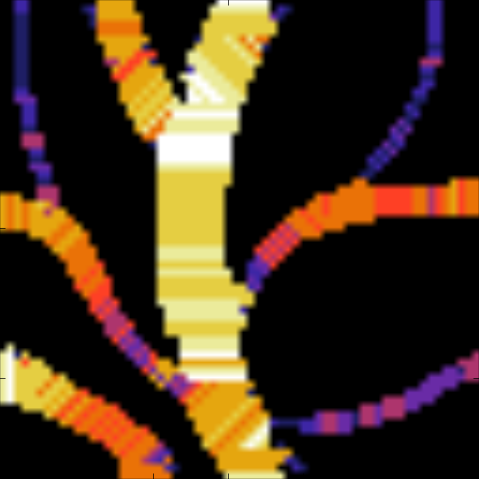} & \includegraphics[scale=0.147]{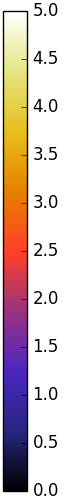}\tabularnewline
\end{tabular}\caption{The color of each pixel in the scale image illustrates the scale of the underlying vessel. Artifacts near the boundaries are removed in a post processing step.
  \label{fig:scale-image}}
\end{figure}

It has been observed previously \cite{descoteaux2008geometric} that
vesselness often fails to correctly estimate the scale of the vessel
along its boundary. This can happen in two cases: When the ridge has a
step-like shape, the curvature near the corner points will be much
larger at the finest scale than at all other scales, leading to an
underestimation of the real scale near the boundary. 
On the other hand, the cross-sectional intensity profile of vessels
may have inflection points near its edges, where $\tilde\nu_2$ changes
its sign. In this case, some points near the boundary will have zero
vesselness at the finest scale, but the sign of $\tilde\nu_2$ will flip,
and therefore vesselness becomes non-zero, at coarser scales, leading
to an overestimation of scale. Figure \ref{fig:scale-image} shows
both scale underestimation or overestimation at vessel boundaries.

While such effects are less problematic for the VED filter, which uses
the same vesselness measure for scale selection, it can lead to
serious artifacts in our filter, where misestimating the scales at
boundaries can cause the curvature-enhancing diffusion to enhance the
boundary of large-scale ridges more than their center.

We avoid such boundary effects by introducing a novel
postprocessing of the computed scales.  For each pixel on a vessel,
the vessel cross-section containing that pixel is extracted by
following the eigenvector direction that corresponds to the strongest
eigenvalue of the Hessian matrix computed at the scale suggested by
the vesselness measures at each point. Then, all pixels are assigned
the scale closest to the average of all pixels that lie on the same
cross-section. This removes the problem of scale over- or
underestimation on the boundaries. Figure \ref{fig:scale-image} shows
the scale image before and after being post processed.
% Another alternative 
%for distributing the vesselness measure to the boundaries of vessels is 
%the method by Descoteaux et al. \cite{descoteaux2008geometric}.
%, where the maximum scale in each
%$3\times3$ neighborhood over the vessel is distributed to each pixel of
%that neighborhood scaled by the projection of the vector --from the center of
%the neighborhood to the considered pixel-- to the cross-section line
%of the vessel at the center.
\subsection{Stability}
\label{sec:l2-stability}
In order to solve Equation~(\ref{eq:ts-fourth-order-diffusion}), we
discretize it with standard finite differences, and use an explicit
numerical scheme. In matrix-vector notation, this can be written as
\begin{equation}
u^{k+1}=u^{k}-\tau Pu^{k}=(I-\tau P)u^{k}\text{ ,} \label{eq:explicit-scheme}
\end{equation}
where $u^{k}\in\mathbb{R}^{m}$ is the vectorized image at iteration
$k$, and the exact form of matrix $P\in\mathbb{R}^{m\times m}$ will be
discussed later. We call a numerical scheme $\ell_{2}$ stable if
\begin{equation}
\left\Vert u^{k+1}\right\Vert _{2}\leq\left\Vert u^{k}\right\Vert _{2}\text{ ,}\label{eq:stability-condition}
\end{equation}
i.e., the $\ell_{2}$ norm of the image is guaranteed not to increase
from iteration $k$ to $k+1$. It follows from
Equation~(\ref{eq:explicit-scheme}) that
\begin{equation}
\left\Vert u^{k+1}\right\Vert _{2}\leq\left\Vert I-\tau P\right\Vert _{2}\cdot\left\Vert u^{k}\right\Vert _{2}\text{ ,}\label{eq:stability-condition-for-our-scheme}
\end{equation}
where $\left\Vert P\right\Vert _{2}$ denotes the $\ell_2$ norm of $P$,
i.e., $\left\Vert P\right\Vert _{2}:=\sqrt{\rho(P^{T}P)}$, where
$\rho(P^T P)$ computes the largest modulus of eigenvalues of the
symmetric matrix $P^T P$.

Consequently, the condition in
Equation~(\ref{eq:stability-condition-for-our-scheme}) is satisfied if
\begin{equation}
  \label{eq:stability-in-terms-of-i-tau-p}
  \left\Vert I-\tau P\right\Vert _{2}\leq 1\text{ .}
\end{equation}
Since $P$ is positive semi-definite, the eigenvalues of $I-\tau P$ are
within the interval
$\left[1-\tau\left\Vert P\right\Vert _{2},1\right]$. Thus,
Equation~(\ref{eq:stability-in-terms-of-i-tau-p}) is satisfied if
$1-\tau\left\Vert P\right\Vert _{2}\geq -1$. This results in the
following constraint on the permissible time step size $\tau$:
\begin{equation}
\label{eq:timestep-bound}
\tau \leq \frac{2}{\left\Vert P\right\Vert_{2}}\text{ .}
\end{equation}

\begin{figure}
\centering%
\begin{tabular}{cc}
\includegraphics[scale=0.082]{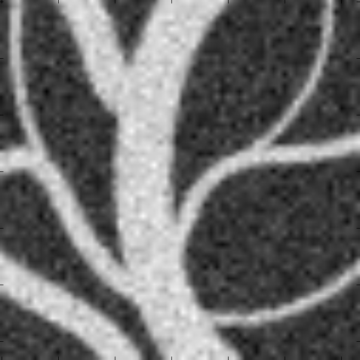} & \includegraphics[scale=0.082]{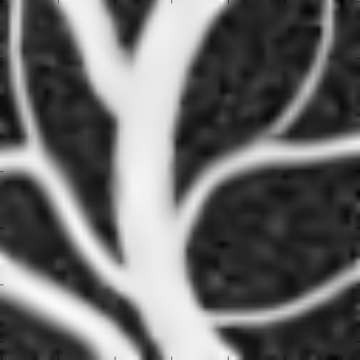}\tabularnewline
$k=0$, $t=0$ & $k=4$, $t=20$\tabularnewline
\includegraphics[scale=0.082]{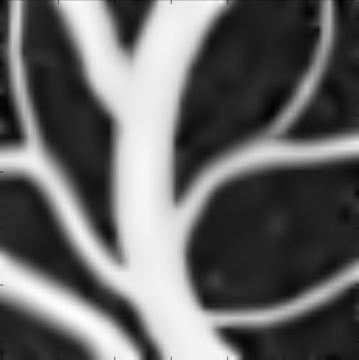} & \includegraphics[scale=0.082]{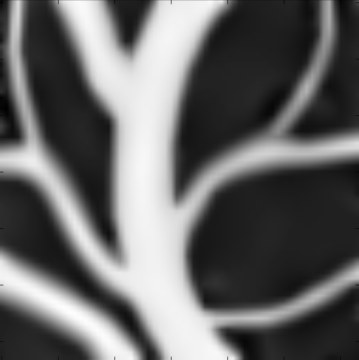}\tabularnewline
$k=8$, $t=40$ & $k=12$, $t=60$\tabularnewline
\end{tabular}
\caption{MAFOD filter applied on the noisy synthesized branch image at different FED cycles $k$. Value $t$ represents the overall diffusion time at each cycle $k$.\label{fig:MAFOD-over-different-cycles}}
\end{figure}

This clarifies that the restriction on the time step size only depends
on $\left\Vert P\right\Vert_{2}$. To compute it, we will now write
down the system matrix $P$ for our discretization of fourth-order
anisotropic diffusion filtering.

Let $L_{xx}$, $L_{xy}$, $L_{yx}$, $L_{yy}$ be matrices approximating
the corresponding derivatives. For ``natural'' boundary condition, it
is important only to approximate the derivatives at pixels $i$ where
the whole stencil fits in the image domain, i.e., where enough data is
available.  Let us combine these four matrices pixelwise into one big
matrix $L$ such that
\begin{equation}
\label{eq:deriv-matrix}
Lu\approx\left(\begin{array}{c}
\vdots\\
\left[L_{xx}u\right]_{i}\\
\left[L_{xy}u\right]_{i}\\
\left[L_{yx}u\right]_{i}\\
\left[L_{yy}u\right]_{i}\\
\vdots
\end{array}\right)\text{,}
\end{equation}
i.e., the approximations of the four derivatives will be next to each
other for every pixel $i$.

The $4\times4$ matrix form of the fourth-order diffusion tensor in
pixel $i$, acting on
\begin{equation*}
\left(\left[L_{xx}u\right]_{i}\text{ }\left[L_{xy}u\right]_{i} \text{ }\left[L_{yx}u\right]_{i}\text{
  }\left[L_{yy}u\right]_{i}\right)^{T},  
\end{equation*}
can be written as $D_i=E M E^{T}$, where $E$ is an orthogonal matrix
containing the vectorized $E_{1}$, $E_{2}$, $E_{3}$, $E_{4}$ from
Equation~(\ref{eq:eigentensors-in-2d}) as its columns and $M$ is a
diagonal matrix with the eigenvalues $\mu_{1}$, $\mu_{2}$, $\mu_{3}$,
$\mu_{4}$ on its diagonal. Due to the choice of the Perona-Malik
diffusivity in our model, $\left\Vert D_i \right\Vert_{2} \leq 1$.

If we arrange all per-pixel matrices $D_i$ in one big matrix $D$ with
a $4\times4$ block-diagonal structure,
\begin{equation}
D=\left(\begin{array}{ccccc}
D_{1} &  & \cdots &  & 0\\
 & D_{2}\\
\vdots &  & D_{3} &  & \vdots\\
 &  &  & \ddots\\
0 &  & \cdots &  & D_{m}
\end{array}\right) \text{,}
\end{equation}
it is clear that $\left\Vert D\right\Vert_{2} \leq 1$, and the whole
scheme reads as
\begin{equation}
u^{k+1}=u^{k}-\tau L^{T}DLu^{k}\text{.} \label{eq:explicit-scheme3}
\end{equation}
Substituting into Equation~(\ref{eq:timestep-bound}) yields
\begin{equation}
\tau \leq \frac{2}{\left\Vert L^{T}DL\right\Vert_{2}}\text{,}
\end{equation}
meaning that, in order to find a stable step size $\tau$, we
have to bound
\begin{equation}
\left\Vert L^{T}DL\right\Vert_{2}\leq
\left\Vert L\right\Vert_{2}^{2} \leq  \sum_{i,j\in\{x,y\}}\left\Vert L_{i,j}\right\Vert _{2}^{2}\text{,}
\end{equation}
whose value will depend on the exact second-order finite difference
stencils. We will use the same discretization as Hajiaboli
\cite{hajiaboli2011anisotropic}, i.e.,
\begin{align*}
u_{xx}&\approx \frac{\left(u_{i-1,j}-2u_{i,j}+u_{i+1,j}\right)}{\left(\Delta x\right)^2}\\
u_{yy}&\approx \frac{\left(u_{i,j-1}-2u_{i,j}+u_{i,j+1}\right)}{\left(\Delta y\right)^2}\\
u_{xy}&\approx \frac{\left(u_{i-1,j-1}+u_{i+1,j+1}-u_{i-1,j+1}-u_{i+1,j-1}\right)}{4\Delta x\Delta y}\\
u_{yx}&= u_{xy}\text{ ,}
\end{align*}
where $\Delta x$ and $\Delta y$ are the pixel edge lengths in $x$ and
$y$ directions, respectively. It is easy to verify using Gershgorin's
theorem that this results in
\begin{equation}
\tau \leq \frac{2}{16\left(\Delta x\right)^2+16\left(\Delta y\right)^2+2\left(\Delta x\Delta y\right)}\text{,}
\end{equation}
i.e., for $\Delta x=\Delta y=1$, $\tau\leq 1/17$.

\subsection{Implementation Using Fast Explicit Diffusion}
\label{sec:fed-scheme}
Since the time step size $\tau$ derived in the previous section is
rather small, solving the discretized version of
Equation~(\ref{eq:ts-fourth-order-diffusion}) numerically using a
simple explicit Euler scheme requires significant computational
effort.
% We are not dealing with higher dimensions in this paper
% Specially in higher dimensions
%the step size must be even smaller and therefore the 
%number of iterations will increase dramatically. 
The recently proposed Fast Explicit Diffusion (FED) provides a
considerable speedup by varying time steps in cycles, in a way that up
to half the time steps within a cycle can violate the stability
criterion, but the cycle as a whole still remains stable
\cite{weickert2016cyclic}. Consequently, a much smaller number of
iterations is required to reach the desired stopping time.

The FED scheme is defined as follows:
\begin{equation}
\begin{array}{cc}
u^{k+1,0}=u^{k}\text{,}\qquad\qquad\qquad\qquad\\
u^{k+1,i+1}=(I-\tau_{i}P(u_{\sigma}^{k}))u^{k+1,i} & \text{ } i=0,\ldots,n-1\text{,}\\
u^{k+1}=u^{k+1,n}\qquad\qquad\qquad\quad\text{ }
\end{array}\label{eq:fed}
\end{equation}
where index $k$ is the cycle iterator, $i$ is the inner cycle
iterator, and $n$ is the number of sub-steps in each cycle.  In order
to ensure stability, $P(u_{\sigma}^{k})$ must be constant during each
cycle. For computing $\tau_i$, first the number of sub-steps in each
cycle must be computed using
\begin{equation}
n=\left\lceil -0.5+0.5\sqrt{1+\frac{12T}{M\tau_{\text{max}}}}\right\rceil \text{,}
\end{equation}
where $T$ is the diffusion stopping time, $M$ is the number of FED cycles
and $\tau_{\text{max}}$ is the step size limit that ensures stability. In our experiments we set $\tau_{max}=0.05$ according to the limit computed in Section \ref{sec:l2-stability}.
As it is shown in \cite{weickert2016cyclic}, $n$ determines $\tau_i$:
\begin{equation}
\tau_{i}=\frac{3T}{2M\left(n^{2}+n\right)\cos^{2}\left(\pi\cdot\frac{2i+1}{4n+2}\right)}\text{, }\left(i=0,\ldots,n-1\right)\text{.}
\end{equation} 
In order to decrease balancing error within each cycle, $\tau_i$'s order should be
rearranged. In our experiments we have used the \emph{$\kappa$-cycles} method for $\tau_i$ reordering \cite{weickert2016cyclic}. 

The fast explicit diffusion framework can be combined with our
discretization in a straightforward manner, and has led to a speedup
of around two orders of magnitude in some of our experiments. For both $\tau_i$ computation and reordering we have used the provided source code by Weickert et al. \cite{weickert2016cyclic}. Figure~\ref{fig:MAFOD-over-different-cycles} shows our filter applied on a synthesized image using the FED scheme with different cycle iterators $k$, corresponding to different stopping times.
%In our experiments on synthetic data, we still used the
%standard Euler scheme with fixed step sizes, as it allows us to find
%the exact stopping time that minimizes the $\ell_{2}$ difference
%between the filtered image and the noise-free ground truth.

\begin{figure*}
\centering%
\begin{tabular}{ccc}
Input & CED & VED \tabularnewline
\includegraphics[scale=0.11]{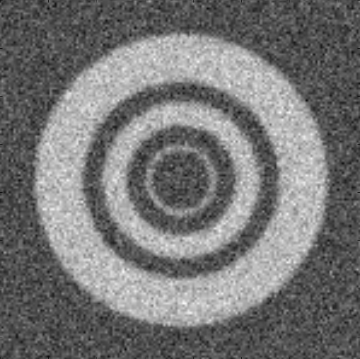} & \includegraphics[scale=0.11]{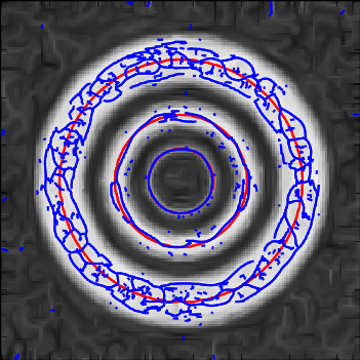} & \includegraphics[scale=0.11]{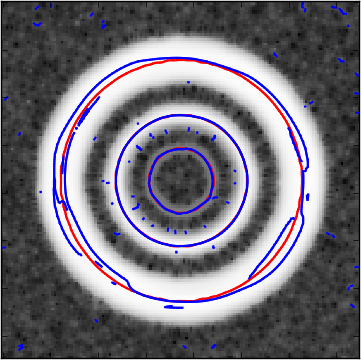}\tabularnewline
& $\mathcal{E}=1.096\text{, }\mathbf{p=100\%}$ & $\mathcal{E}=1.116\text{, }\mathbf{p=100\%}$\\[1ex]
IFOD & Single-scale Gaussian & Multi-scale Gaussian \tabularnewline
\includegraphics[scale=0.11]{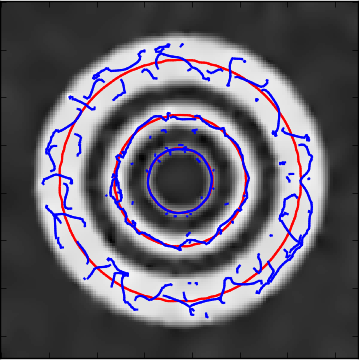} & \includegraphics[scale=0.11]{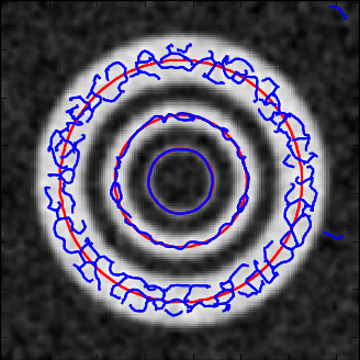} & \includegraphics[scale=0.11]{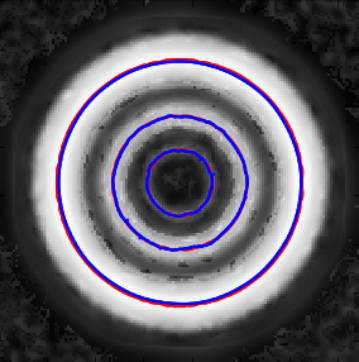}\tabularnewline
 $\mathcal{E}=1.579\text{, }p=96\%$ & $\mathcal{E}=1.263\text{, }p=98\%$ & $\mathcal{E}=0.566\text{, }\mathbf{p=100\%}$ \\[1ex]
 Bilateral & Hajiaboli & MAFOD \tabularnewline
 \includegraphics[scale=0.11]{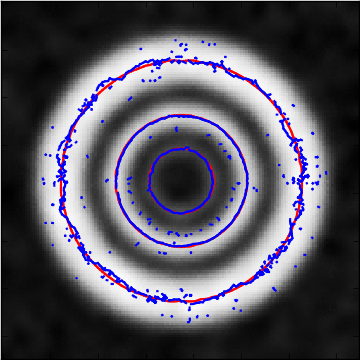} & \includegraphics[scale=0.11]{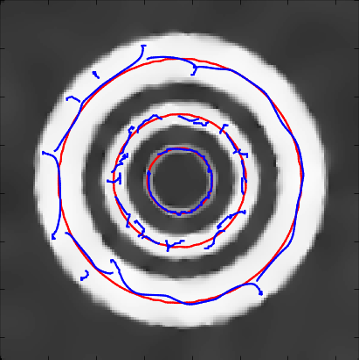} & \includegraphics[scale=0.11]{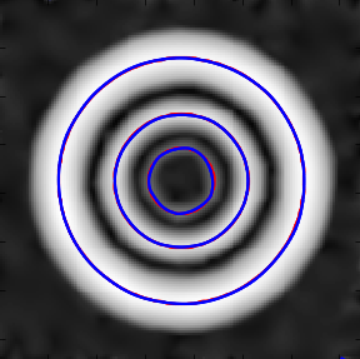} \tabularnewline
  $\mathcal{E}=0.624\text{, }p=99\%$ & $\mathcal{E}=1.024\text{, }p=92\%$ & $\mathbf{\mathcal{E}=0.332\text{, }p=100\%}$  \tabularnewline
\end{tabular}\caption{Red curves show the ground truth ridge location, while blue curves 
show the location reconstructed from the filtered noisy image. Our
MAFOD filter restores ridge locations from the noisy image with ridges of different 
scales better than other filters. 
% Image size is $150\times150$ pixels.
\label{fig:concentric-circles}}
\end{figure*}

\begin{figure*}
%\centering
\floatbox[{\capbeside\thisfloatsetup{capbesideposition={left,top},capbesidewidth=4cm}}]{figure}[\FBwidth]
{\caption{Each plot shows the intensities on a left to center line of the 2D images in  Figure~\ref{fig:concentric-circles}. The lines are taken from the middle of each image. The red dashed lines show the true position of the ridge points, and the blue lines show the position of local maxima over the intensity line scan of each filtered image. }\label{fig:concentric-circles-line-scan}}
{\begin{tabular}{ccc}
Input & CED & VED\tabularnewline
\includegraphics[scale=0.52]{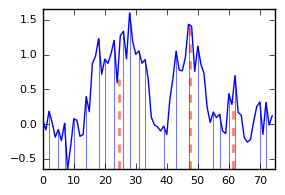} & \includegraphics[scale=0.52]{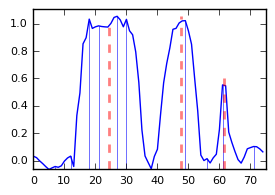} & \includegraphics[scale=0.52]{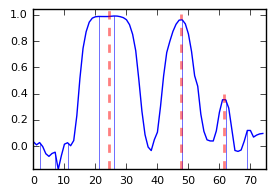}\tabularnewline
IFOD & Single-scale Gaussian & Multi-scale Gaussian\tabularnewline
\includegraphics[scale=0.52]{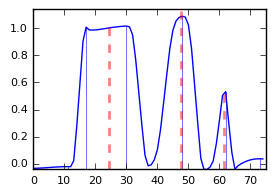} & \includegraphics[scale=0.52]{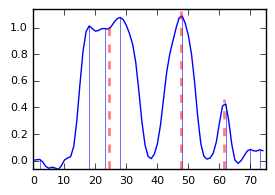} & \includegraphics[scale=0.52]{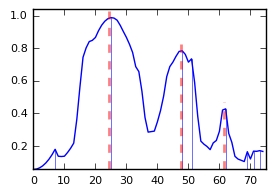}\tabularnewline
Bilateral & Hajiaboli & MAFOD\tabularnewline
\includegraphics[scale=0.52]{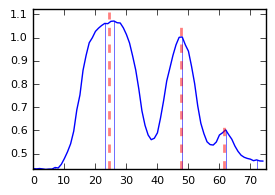} & \includegraphics[scale=0.52]{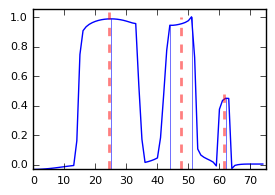} & \includegraphics[scale=0.52]{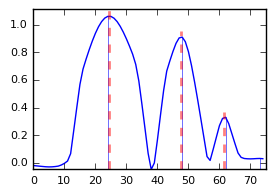}\tabularnewline
\end{tabular}}

\end{figure*}

\subsection{Ridge and Valley Extraction}
\label{sec:ridge-detection-scale-selection}

After enhancing ridges and valleys with our filter, we extract a
polygonal representation of them using a 2D counterpart of an
established 3D algorithm \cite{schultz2010crease}. Our algorithm is
based on the idea of marching squares \cite{schroeder2005overview} and
involves the zero contour of the scalar field $d=\text{det}(g|Hg)$,
where $(g|Hg)$ indicates a matrix whose first column is the local
gradient vector $g$ and the second column $Hg$ is the result of
multiplying the gradient vector to the Hessian matrix;
$\text{det}(\centerdot)$ is the matrix determinant. The zero level set
of $d$ is a superset of the creases \cite{peikert2008height}.

Our overall approach of ridge extraction involves two different
notions of scale: The first one refers to the selection of scales at
which derivatives are taken, as discussed in
Section~\ref{sec:scale-selection}; the second one to the stopping time
$t$ of our filter. To clarify their respective roles, we compare our
approach to the seminal work by Lindeberg \cite{lindeberg1998edge} on
ridge extraction in Gaussian scale space.

In Lindeberg's approach, ridge curves in 2D images sweep out surfaces
in three-dimensional scale space, and curves on these surfaces are
found along which a measure of ridge strength is locally maximal with
respect to diffusion time $t$. An example of such a measure is
\begin{equation}
R(u_{\sigma})=t^{4\gamma}\left(u_{xx}+u_{yy}\right)^{2}\left(\left(u_{xx}-u_{yy}\right)^{2}+4u_{xy}^{2}\right)\text{.}
\label{eq:ridge-strength}
\end{equation}
In this approach, stopping time $t$ and the scale $\sigma$ of
derivatives are related by $t=\sigma^2/2$, and can thus be considered
as one single parameter, whose value is determined automatically. The
exponent $\gamma$ in the normalization factor that is used to
compensate for the loss of contrast at later diffusion times is
treated as a tunable parameter. In our experiments, we set it to
$\gamma=\frac{3}{4}$, as proposed in \cite{lindeberg1998edge}.

Decoupling the $t$ and $\sigma$ parameters is a price that we pay in
our method in order to preserve and enhance creases, for which
Gaussian scale space does not have any mechanism. Our current
implementation selects the derivative scale $\sigma$ automatically, as
discussed in Section~\ref{sec:scale-selection}, but does not have an
objective criterion for setting the stopping time $t$, unless ground
truth is available. In practice, we found it relatively simple to tune
this parameter based on viewing the corresponding images, especially
given that, after image noise has been removed, results are relatively
stable (cf.\ Figure~\ref{fig:MAFOD-over-different-cycles}). Future
work might investigate automated selection of this parameter.

Another difference between our approach and Lindeberg's is that his
crease extraction algorithm operates on the full scale space, while
ours, similar to previous work by Barakat et al. \cite{Barakat:2011},
works on a single, pre-filtered image. Both approaches have relative
benefits and drawbacks: Scale space crease extraction is challenging
to implement, and requires much more time and memory, especially when
dealing with the four-dimensional scale space resulting from
three-di\-men\-sion\-al input images \cite{Kindlmann:2009Vis}. On the other
hand, it might, in rare cases, indicate spatially intersecting creases
at different scales, which our current approach is not able to
reproduce.

\section{Experimental Results}
\label{sec:experimental-results}

We compare our multi-scale anisotropic fourth-order diffusion (MAFOD)
to crease enhancement diffusion (CED) \cite{sole2001crease},
vesselness enhancement diffusion (VED) \cite{canero2003vesselness},
isotropic fourth-order diffusion (IFOD) \cite{lysaker2003noise}, the
anisotropic fourth-order diffusion by Hajiaboli
\cite{hajiaboli2011anisotropic}, bilateral, and a multi-scale Gaussian
filter. Since it was already shown in  \cite{sole2001crease} that the 
coherence enhancing diffusion filter \cite{weickert1998anisotropic} 
tends to more strongly deform non linear structures compared to the CED 
filter, it is not included in the comparison.

The multi-scale Gaussian filter is defined to approximate Lindeberg's
scale selection, as described in
Section~\ref{sec:ridge-detection-scale-selection}. From a range of
stopping times between $t=1$ and $t=30$, it first selects an optimal
scale for each pixel, by finding the $t$ that maximizes $R(u_\sigma)$
from Equation~(\ref{eq:ridge-strength}). Then, the intensity of each
pixel in the output image is obtained by convolving the input image
with a Gaussian at the locally optimal scale $\sigma=\sqrt{2t}$ that
is then normalized between $\left[0,1\right]$. The normalization is
necessary to compensate for the intensity range shrinkage after
Gaussian blurring.

The crease extraction algorithm from
Section~\ref{sec:ridge-detection-scale-selection} results in a set of
polygonal chains. For each crease line segment in the ground truth, a
corresponding segment in the reconstruction is selected by picking the
one with minimum Hausdorff distance \cite{huttenlocher1993comparing}
in a neighborhood around the ground truth line segment.  This
neighborhood is set to six pixels for the experiments on synthetic
data, and to ten pixels for real data.  The average
Euclidean distance $\mathcal{E}$ between the ground truth and the
corresponding reconstruction is then used to quantify the accuracy of
vessel locations in the filtered image. In addition to $\mathcal{E}$,
we show the percentage $p$ of ground truth for which a
corresponding ridge was detected from the filtered images while
computing $\mathcal{E}$.

In the experiments on synthesized images, image evolution of all
filters, except for multi-scale Gaussian and bilateral filters, was
stopped when the $\ell_{2}$ difference between the filtered image and
the noise-free ground truth was minimized. $\ell_2$ difference was
chosen over $\mathcal{E}$ as a stopping criterion due to its much
lower computational cost.

\subsection{Confirming Theoretical Properties}
\label{sec:simple-synthetic-image}

%\begin{figure}[t]
%\centering\includegraphics[scale=0.35]{Images/concentric-circles}\protect\caption{
%Red curves show the vessel location of ground truth and blue curves show the
%vessel location of the filtered noisy image. Our
%MAFOD filter restores ridge locations from the noisy image with ridges of different 
%scales better than other filters.\label{fig:concentric-circles}}
%\end{figure}

% Weickert's second order coherence enhancing filter 
% \cite{weickert1995multiscale}  is not in the comparison 
% list as it destroys junctions and nonlinear structures 
% \cite{sole2001crease}. 

Our filter has been designed to improve localization accuracy while
accounting for creases at multiple scales and being rotationally
invariant. Results on a simple simulated image with three concentric
ridges of different radii, which is contaminated with zero-mean
Gaussian noise with a signal to noise ratio $\mathrm{SNR}=6.81$,
verify that these design goals are met.

Both Figure~\ref{fig:concentric-circles} and Figure~\ref{fig:concentric-circles-line-scan} show that our MAFOD filter restores ridge
locations most accurately as assessed both by visual inspection and
Euclidean distance $\mathcal{E}$. MAFOD outperforms CED, IFOD,
Hajiaboli and bilateral filtering since it accounts for different
scales. On the other hand, the curvature enhancement of our filter,
which is not part of multiscale VED or Gaussian filters, clearly makes
it easier for the ridge extraction algorithm to localize the
centerline, especially in the largest ridge. IFOD does perform
curvature enhancement but, due to its isotropic nature, it is not
effectively guided to act specifically across the ridge. As it is obvious 
on the largest circle, the multi-scale Gaussian filter leads to ridge displacement.
The result of the anisotropic fourth-order filter by Hajiaboli clearly
illustrates the fact that it was designed to preserve edges, not to
enhance creases.

\begin{figure*}
\centering%
\begin{tabular}{ccc}
Ground Truth & CED & VED\tabularnewline
\includegraphics[scale=0.092]{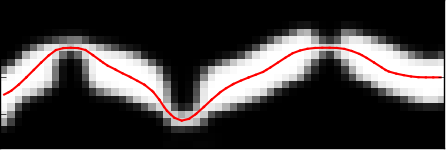} & \includegraphics[scale=0.092]{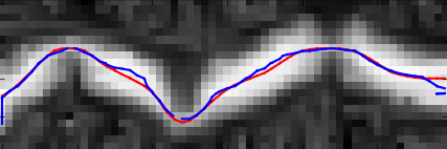} & \includegraphics[scale=0.092]{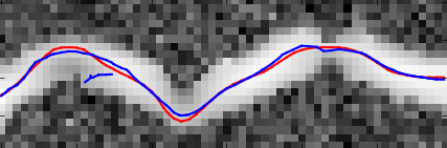}\tabularnewline
 & $\mathcal{E}=0.343\text{, }p=98\%$ & $\mathcal{E}=0.345\text{, }\mathbf{p=100\%}$\tabularnewline
Multi-scale Gaussian & Bilateral & MAFOD\tabularnewline
\includegraphics[scale=0.092]{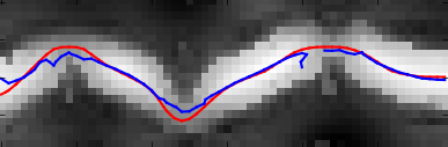} & \includegraphics[scale=0.092]{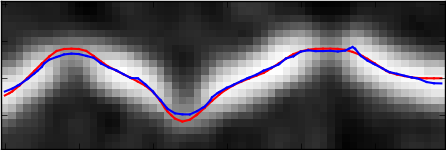} & \includegraphics[scale=0.092]{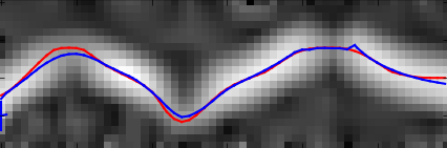}\tabularnewline
$\mathcal{E}=0.506\text{, }p=97\%$ & $\mathcal{E}=0.341\text{, }\mathbf{p=100\%}$ & $\mathbf{\mathcal{E}=0.302\text{, }p=100\%}$\tabularnewline
\end{tabular}\caption{In
  a simulated occluded vessel, restored (blue) curves again best match
  the (red) ground truth locations in case of our MAFOD filter. In
  addition, MAFOD better preserves the occlusions than VED. 
  % image with size $20\times60$ pixels.
  \label{fig:vessel-occlusion}}
\end{figure*}

For the MAFOD filter, scales $\sigma$ and
vesselness threshold $\theta$ are the same as for VED, 
$\sigma=\{0.5,1.0,...,8.5,9.0\}$, $\theta=0.2$. Other parameters 
are $\lambda=0.005$ for MAFOD, IFOD, and $\sigma=1.0$
for IFOD; for Hajiaboli, $\lambda=0.01$; for CED, $\sigma=2.0$ and it is
set to enhance both ridges and valleys; for single-scale Gaussian smoothing,
$\sigma=1.25$. For the MAFOD filter, FED stopping time is set
to $500$, and the number of cycles is set to $10000$. 
For other fourth-order equations $\tau=0.03$ and for the second-order 
diffusion equations such as the VED and CED filters, $\tau=0.2$; for the
bilateral filter $\sigma_{spatial}=3.0$ and $\sigma_{range}=1.0$. 

% Has this be chosen systematically from a range of potential \sigmas
% to minimize the L2 error? If so, we don't have to mention it
% For the bilateral filter,
%$\sigma_{spatial}=5.0$ and $\sigma_{range}=0.9$.

% I am omitting \tau because it was never introduced. Also,
% if chosen small enough, it should not have a big
% effect on the result, just on the computational effort
% $\tau=0.03$ for all

% Moreover enhancing either valleys or ridges
% reduces the contrast of the filtered image therefore for the next
% experiments both ridges and valleys are enhanced.

\subsection{Simulated Vessel Occlusion}
\label{sec:simulated-vessel}

Figure~\ref{fig:vessel-occlusion} shows a second image, simulating an
occluded vessel, and corrupted with Gaussian noise with
$\mathrm{SNR}=6.40$. Our MAFOD filter leads to the most accurate
localization in terms of Euclidean error $\mathcal{E}$. In particular,
we observed that VED widens the occlusions. They are better preserved
by our filter, which we set to enhance both ridges and valleys.

Again, an amount of smoothing that minimized $\ell_2$ error was used for
all filters except for multi-scale Gaussian and bilateral filters.  
The parameters for VED and MAFOD
are $\sigma=\{0.5,1.0,1.5,2.0,2.5,3.0\}$, and $\theta=0.35$;
$\lambda=0.017$, stopping time is $20$ and
the number of cycles is set to $1000$ for MAFOD; 
for CED, $\sigma=1.0$; for the bilateral filter,
$\sigma_\text{spatial}=1.5$ and $\sigma_\text{range}=1.0$. 
For the numerical solver, we set $\tau=0.05$
for the fourth-order equation and for second-order equations, $\tau=0.2$.

%\begin{figure}[t]
%\centering\includegraphics[scale=0.295]{Images/vessel-occlusion}\protect\caption{In
%  a simulated occluded vessel, restored (blue) curves again best match
%  the (red) ground truth locations in case of our MAFOD filter. In
%  addition, MAFOD better preserves the occlusions than VED.\label{fig:vessel-occlusion}}
%\end{figure}

\subsection{Real Vessel Tree}
\label{sec:real-vessel-tree}

\begin{figure*}[t]
\centering%
\begin{tabular}{ccccc}

 Input & VED filter & Bilateral Filter & Ms Gaussian & MAFOD filter\tabularnewline
\includegraphics[scale=0.58]{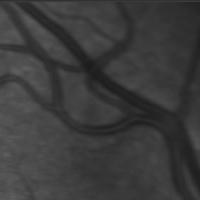} & \includegraphics[scale=0.08]{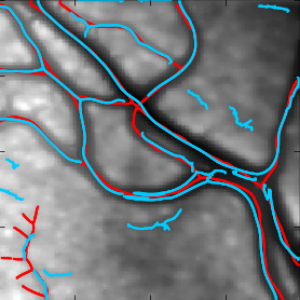} & \includegraphics[scale=0.08]{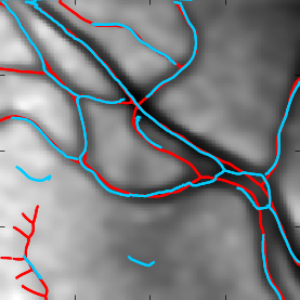} & \includegraphics[scale=0.08]{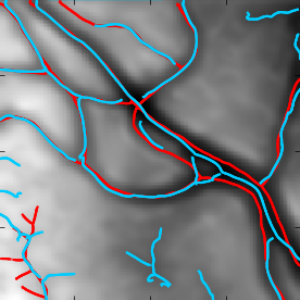} & \includegraphics[scale=0.08]{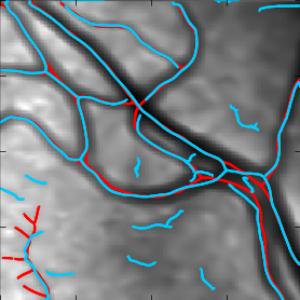}\tabularnewline
ROI 1 & $\mathcal{E}=1.45$ & $\mathcal{E}=1.36$ & $\mathcal{E}=1.89$ & $\mathbf{\mathcal{E}=1.17}$\tabularnewline

\includegraphics[scale=0.58]{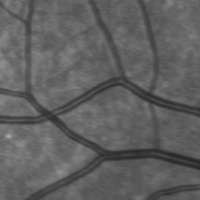} & \includegraphics[scale=0.29]{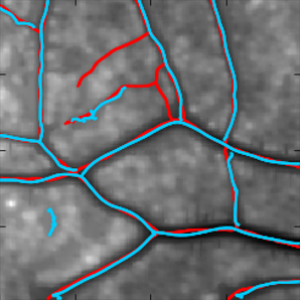} & \includegraphics[scale=0.29]{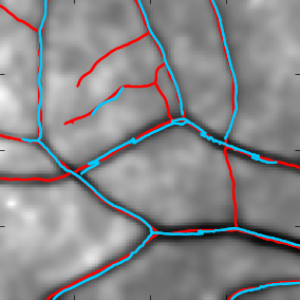} & \includegraphics[scale=0.29]{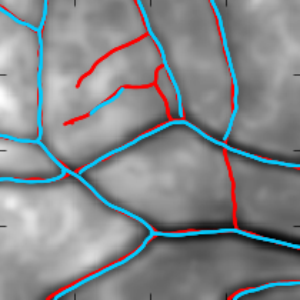} & \includegraphics[scale=0.29]{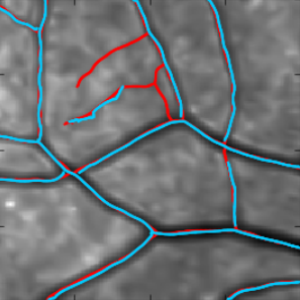}\tabularnewline
ROI 2 & $\mathcal{E}=1.31$ & $\mathcal{E}=1.42$ & $\mathcal{E}=1.27$ & $\mathbf{\mathcal{E}=1.04}$\tabularnewline

\includegraphics[scale=0.58]{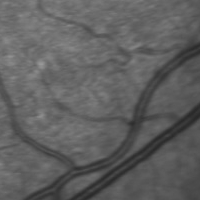} & \includegraphics[scale=0.29]{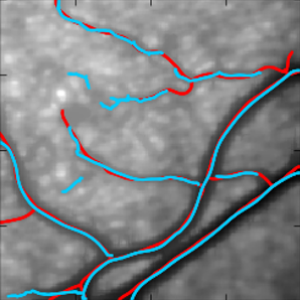} & \includegraphics[scale=0.29]{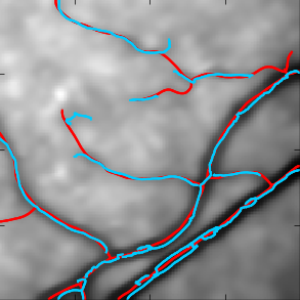} & \includegraphics[scale=0.29]{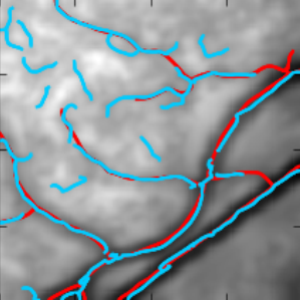} & \includegraphics[scale=0.29]{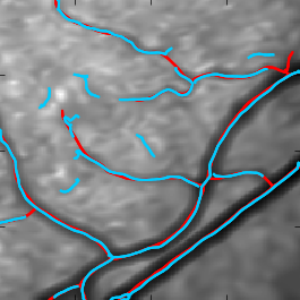}\tabularnewline
ROI 3 & $\mathcal{E}=1.43$ & $\mathcal{E}=1.63$ & $\mathcal{E}=1.62$ & $\mathbf{\mathcal{E}=1.15}$\tabularnewline
      & $p=89\%$ & $p=80\%$ & $p=88\%$ & $\mathbf{p=92\%}$\tabularnewline
  
\end{tabular}\protect\caption{In three ROIs of a fundus image,
  reconstructed vessel locations (blue) best match a manually marked
  ground truth (red) when our MAFOD filter is used. 
  \label{fig:real-data}}
\end{figure*}

To demonstrate our filter on a real-world example, we applied it to
several ROIs from an infrared fundus image, on which one of our co-authors
(MWMW), who is an ophthalmologist, manually marked the exact vessel
locations to provide a ground truth for comparison, without being
shown the filtered images. Results in Figure~\ref{fig:real-data} show
that our MAFOD filter outperforms VED, multi-scale Gaussian and
bilateral filters in restoring vessel locations. In ROI~3, at some
point the two thickest vessels run close to each other. By looking at
the filtered image with the VED, the two vessels are erroneously
connected to each other in that area, even though they are not connected in
the corresponding extracted valley curves. Our MAFOD filter
correctly avoided connecting the vessels to each other.

Even though vessels generally appear dark (i.e., as valleys) in these
images, the larger ones exhibit a thin ridge at their center, due to a
reflex in the infrared image. This leads to an incorrect double
response in single-scale filters as shown for the bilateral
filter. CED and IFOD filters suffer from similar problems (results not
shown).

For each filter separately, we carefully tuned the parameters for
optimum results. Specially $\theta$, $\lambda$ and the stopping time
are the parameters that need more careful tuning compared to others. 
For the MAFOD filter, we set
$\sigma=\{0.2,0.3,0.5,1.0,...,6.5,7.0\}$, $\lambda=0.005$,
$\theta=0.13$, and used a FED scheme with stopping time $12$ and cycle
number $2$.  For the VED filter, an
explicit Euler scheme is used with $600$ iterations and $\tau=0.2$,
and the same parameters for scale selection as for MAFOD; for the
bilateral filter, $\sigma_\text{range}=0.3$ and
$\sigma_\text{spatial}=3.0$; for the multi-scale Gaussian filter 
an additional Gaussian smoothing with kernel size $\sigma=2$ is
applied to the filtered image to blur out discontinuities from scale
selection and thus achieve an even better result.  The computational
effort of all filters is reported in Table~\ref{tab:fundus-time}.

\begin{table}
\centering%
\caption{The average filtering time for a single ROI of size $200\times200$ pixels 
in Figure~ \ref{fig:real-data}.\label{tab:fundus-time}}
\label{tab:1}       % Give a unique label
\begin{tabular}{lllll}
\hline\noalign{\smallskip}
Filter & VED & Bilateral & Ms Gaussian &  MAFOD \\
\noalign{\smallskip}\hline\noalign{\smallskip}
Time ($sec$) & $1251$ & $0.16$ & $2.24$ & $6.51$ \\

\noalign{\smallskip}\hline
\end{tabular}
\end{table}

\section{Conclusion}
\label{sec:conclusion}

We have proposed a new multi-scale fourth order anisotropic diffusion (MAFOD) filter to enhance ridges and valleys in images. It uses a fourth order diffusion tensor which smoothes along creases, but sharpens them in the perpendicular direction, and optionally enables enhancing either ridges or valleys only. Our results indicate that the curvature enhancing properties of fourth-order diffusion allow our filter to better restore the exact crease locations than traditional methods. In addition, we found that our filter better preserves vessel occlusions.

In the future, we would like to extend our 2-D filter to 3-D images,
and to better handle crossings and bifurcations
\cite{hannink2014crossing}.

% Crease lines are useful tools for extraction of object cores or skeletal
% structures in image processing and computer vision. Particularly in
% medical images, vessels are the crease lines and thus they can be
% enhanced or segmented using crease enhancement or crease segmentation
% methods.

%\begin{acknowledgements}
%\end{acknowledgements}


\bibliographystyle{abbrv}
\bibliography{literature}

%\begin{thebibliography}{}
% etc
%\end{thebibliography}

\end{document}